# Lane Departure Prediction Based on Closed-Loop Vehicle Dynamics

Daofei Li*[1], Siyuan Lin[1] and Guanming Liu[1]

**Abstract**

An automated driving system should have the ability to supervise its own performance and to request human driver to take over when necessary. In the lane keeping scenario, the prediction of vehicle future trajectory is the key to realize safe and trustworthy driving automation. Previous studies on vehicle trajectory prediction mainly fall into two categories, i.e. physics-based and manoeuvre-based methods. Using a physics-based methodology, this paper proposes a lane departure prediction algorithm based on closed-loop vehicle dynamics model. We use extended Kalman filter to estimate the current vehicle states based on sensing module outputs. Then a Kalman Predictor with actual lane keeping control law is used to predict steering actions and vehicle states in the future. A lane departure assessment module evaluates the probabilistic distribution of vehicle corner positions and decides whether to initiate a human takeover request. The prediction algorithm is capable to describe the stochastic characteristics of future vehicle pose, which is preliminarily proved in simulated tests. Finally, the on-road tests at speeds of 15 to 50 km/h further show that the proposed method can accurately predict vehicle future trajectory. It may work as a promising solution to lane departure risk assessment for automated lane keeping functions.

## 1. Introduction

Driving in the expected lane in a safe and comfortable fashion is one of the core tasks of Advanced Driving Assistant Systems such as LKA (Lane Keeping Assist) and LCA (Lane Centering Assist). When such systems are not capable to handle the dynamic driving tasks, a risk assessment module is needed to provide timely warning to the human driver and to initiate necessary human takeover requests. For highly-automated driving systems, e.g. level 3 conditional driving automation per SAE J3016 standard, such risk assessment module becomes extremely crucial both for safety and customer experience. Inaccurate risk assessment algorithms, aggressive or conservative, will discourage the use of such automated driving functions. For example, if a risk assessment algorithm cannot precisely judge whether an LKA system can handle lane keeping tasks or not, but initiates takeover requests too frequently, the user experience is significantly compromised. Therefore, there are urgent needs for an accurate algorithm of risk assessment, which as a watchdog algorithm can execute online supervision over the closed-loop performances of automated driving.

A successful risk assessment algorithm should be based on accurate perception or prediction of surrounding road environment, including that of other traffic users and lane boundaries. Then, the ego vehicle motion prediction algorithm is to predict the future positions and poses of vehicle body, with which a risk level is obtained for the final assessment of a human takeover request.

As the core of risk assessment and also the focus topic of this paper, the ego vehicle motion prediction is comprised of three parts, i.e. vehicle state estimation via sensor fusion, motion prediction and takeover request assessment. As for state estimation, there have been extensive studies using Kalman Filter (KF) or extended Kalman Filter (EKF) to compensate sensor errors[1,2], or Unscented Kalman Filter (UKF) for highly nonlinear systems[3,4,5,6].

As for vehicle motion prediction, previous studies can be mainly divided into the physics-based and the manoeuvre-based approaches[7]. The physics-based motion models, as the name indicates, are based on understanding of vehicle physics, in kinematic or dynamic ways. Popular kinematic models include the Constant Velocity (CV) model, the Constant Acceleration (CA) model, the Constant Turn Rate and Velocity (CTRV) model and the Constant Turn Rate and Acceleration (CTRA) model[1,2,3,4,8], while the dynamic models include the 2 or 3 degrees of freedom (DOFs) 'bicycle' model[9,10]. Though with great computing efficiency, existing algorithms with physics-based approach cannot consider the control operation from driver or automation agent, so the long-term prediction accuracy cannot be guaranteed.

The second category of vehicle motion prediction, the manoeuvre-based approach, treats driving as a series of manoeuvres and requires more machine learning from obtained datasets. Several machine learning approaches have been used as classifiers to identify the driving intention of the target vehicle, e.g. HMM (Hidden Markov Model)[11,12], DBN (Dynamic Bayesian Network)[5,13], and SVM (Support Vector Machine)[14,15]. For end-to-end prediction of vehicle trajectory, neural networks have been often adopted, e.g. LSTM (Long Short Term Memory)[16,17,18,19], NARNN (Nonlinear Autoregressive Neural Network) and FFNN (Feed-Forward Neural Network)[15], DFNN (Deep Fourier

[1] Institute of Power Machinery and Vehicular Engineering, College of Energy Engineering, Zhejiang University.

* Corresponding author: Daofei Li, Institute of Power Machinery and Vehicular Engineering, College of Energy Engineering, Zhejiang University, No 38 Zheda Road, Xihu District, Hangzhou, 310028, China. Email: dfli@zju.edu.cn

Neural Network)[20], CoverNet[21], GMM (Gaussian Mixture Model)-based MDN (Mixture Density Network)[22]. Given sufficient amount and scenario coverage of training data, the manoeuvre-based models can identify and predict the driver's intended operation and thus ensure the long-term prediction accuracy. However, this excellent performance in the theoretical sense calls for a great deal of thorough validation, particularly to meet SOTIF (Safety Of The Intended Functionality) requirements.

By comparing these two approaches, we find that for ego vehicle motion prediction in automated driving, it is better to extend the physics-based approach, especially when considering the better availability of vehicle states and the convenience of SOTIF validation. Furthermore, in an automated vehicle, its control actions are generated according to tracking control law, so basically it can be easier to predict the future manoeuvres as long as the control law is known. This can be viewed as an improved version of physics-based approach, which takes advantage of the closed-loop vehicle dynamics.

In this paper, we focus on lane keeping scenarios of automated driving vehicles as shown in Figure 1. The lane centreline is chosen as the target trajectory, while the left and right lane lines hereinafter can be the physical lane marker lines of road or the imaginary lane lines according to different lane keeping preferences. The rectangle, which is slightly larger than the actual vehicle, is used to indicate the vehicle planar contour. The task is to get the predicted vehicle pose (in dashed green) based on the current vehicle pose (in solid black), and to quantify the risk of lane departure at future time steps. For the front left corner, the light green area schemes its possible position at the current step, while the uncertainties are due to the fusion errors of observers and sensors. The dark green area schemes its possible position at the predicted future time step, while *d* is its marginal distance from the left lane line.

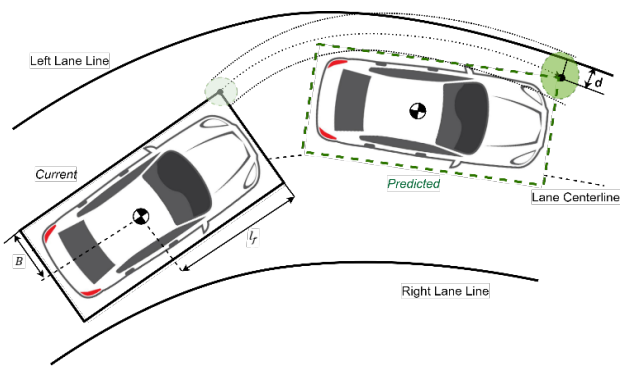

Figure 1 Vehicle prediction in lane keeping senario

To fulfil this, a lane departure prediction method based on closed-loop vehicle dynamics model is proposed, with the framework shown in Figure 2. Unlike the existing approaches of trajectory prediction, here we do not use only sensing and planning modules, but also tracking control module in designing the algorithm. As a part of the entire automated driving system, the necessary laws or strategies for control prediction can be directly migrated from the control module. If the module does not perform as a white-box to the prediction algorithm developer, the control laws can be learned via controller-in-the-loop simulation or on-road tests.

The main parts include extended Kalman Filter (EKF), Kalman Predictor with Control (KPC), and Lane Departure Assessment (LDA). EKF is to estimate the current vehicle states based on sensing module outputs. With known tracking control laws, KPC is used to obtain the control inputs and to predict vehicle states in the future. Finally, LDA evaluates the risk of lane departure and outputs a departure flag for takeover decision.

The rest of the paper is organized as follow. In Section 2, a single-track vehicle model is established for lane keeping scenarios. Section 3 details the vehicle prediction algorithm, including tracking control, EKF and KPC. Section 4 presents risk assessment of lane departure. Then the proposed algorithm is verified by both simulation and road tests, in Sections 5 and 6, respectively. Finally, conclusions and future work are given in Section 7.

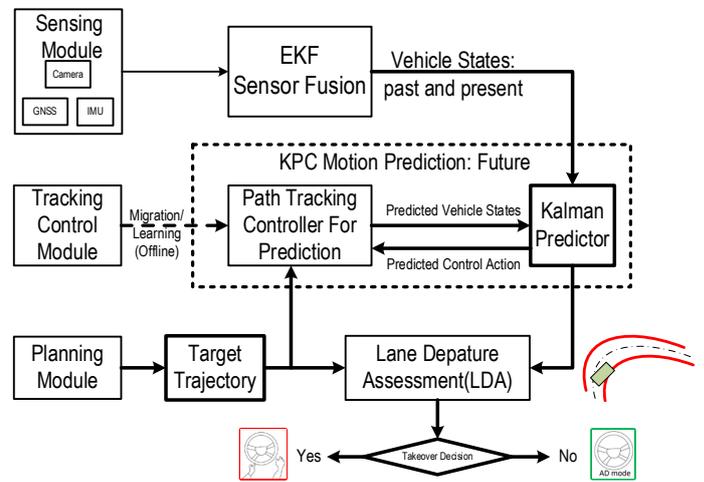

Figure 2 Proposed framework of lane departure prediction

## 2. Vehicle Dynamics Modelling

A single-track model with 2 DOFs, i.e. the lateral and yaw motion, is adopted[23]. Here we assume the vehicle runs on the flat road surface with a constant longitudinal speed, and ignore both the vehicle pitch and roll motion. Figure 3 shows the simplified model in the moving coordinate system xoy, and also the inertia coordinate system XOY.

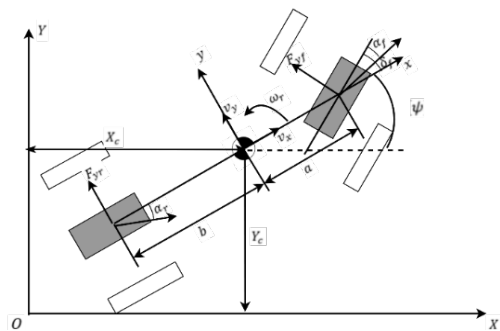

Figure 3 Single-track vehicle dynamics model

The following equations can be obtained according to Newton's Second Law.

$$\begin{cases} \dot{v}_y = \Sigma F_y/m - v_x\omega_r \\ \quad = (F_{yf}\cos(\delta_f) + F_{yr})/m - v_x\omega_r \\ \dot{\omega}_r = \Sigma M_z/I_z = (aF_{yf}\cos(\delta_f) - bF_{yr})/I_z \end{cases} \quad (1)$$

where $v_x$ denotes the constant longitudinal velocity of vehicle centre of gravity (CG); $v_y$ denotes the CG lateral velocity; $\omega_r$ denotes the vehicle yaw rate; $F_{yf}$ and $F_{yr}$ denote the lateral forces of front and rear tyres, respectively; $\alpha_f$ and $\alpha_r$ denote the slip angles of front and rear tyres, respectively; $\delta_f$ denotes the front wheel steering angle; $m$ and $I_z$ denote the mass and yaw moment of inertia of the vehicle, respectively; $a$ and $b$ denote the distances from the CG to the front and rear axles, respectively. All symbols and vehicle parameters are listed in Appendix 1.

Assuming that in the lane keeping process, the vehicle lateral acceleration is less than 0.4g, so the tyre lateral force characteristics are in the linear range, and the influence of load transfer can also be ignored. Then we have

$$\begin{cases} F_{yf} = -C_f\alpha_f = -C_f((v_y + a\cdot\omega_r)/v_x - \delta_f) \\ F_{yr} = -C_r\alpha_r = -C_r(v_y - b\cdot\omega_r)/v_x \end{cases} \quad (2)$$

where $C_f$ and $C_r$ are the total cornering stiffnesses of the front and rear tyres, respectively.

In the inertia coordinate system XOY, the vehicle motion equations can be expressed as,

$$\begin{cases} \dot{X}_c = v_x\cos(\psi) - v_y\sin(\psi) \\ \dot{Y}_c = v_x\sin(\psi) + v_y\cos(\psi) \\ \dot{\psi} = \omega_r \end{cases} \quad (3)$$

where $X_c$ and $Y_c$ denote the abscissa and ordinate of the vehicle's CG in the inertial coordinate system; $\psi$ denotes the heading angle of vehicle.

The system states are selected as $[v_y\ \omega_r\ X_c\ Y_c\ \psi]^T$, and the control input is front wheel steering angle $\delta_f$. Then by combining equations (1-3), the vehicle motion in lane keeping process can be described as

$$\begin{cases} \dot{v}_y = -\dfrac{C_f + C_r}{mv_x}v_y - \left(\dfrac{aC_f - bC_r}{mv_x} + v_x\right)\omega_r + \dfrac{C_f}{m}\delta_f \\ \dot{\omega}_r = -\dfrac{aC_f - bC_r}{I_zv_x}\cdot v_y - \dfrac{a^2C_f + b^2C_r}{I_zv_x}\omega_r + \dfrac{aC_f}{I_z}\delta_f \\ \dot{X}_c = v_x\cos(\psi) - v_y\sin(\psi) \\ \dot{Y}_c = v_x\sin(\psi) + v_y\cos(\psi) \\ \dot{\psi} = \omega_r \end{cases} \quad (4)$$

To identify the system parameters, we carry out double-lane-change tests at two speeds of 30 km/h and 50 km/h. The identified model is simulated with the exact steering inputs from road tests. Figure 4 presents the steering angle $\delta_f$ and vehicle responses of both simulation ("Sim") and test measurements ("Mea"), including lateral velocity $v_y$, vehicle trajectories (North-East), yaw rate $\omega_r$, heading angle $\psi$ and lateral acceleration $a_y$. The results show that the identified model can accurately describe the vehicle lateral and yaw motion in the considered speed range. In real applications, to account for the variation of vehicle parameters and the effects of unknown disturbances from environment, an online estimator can be designed to guarantee the accuracy of vehicle model.

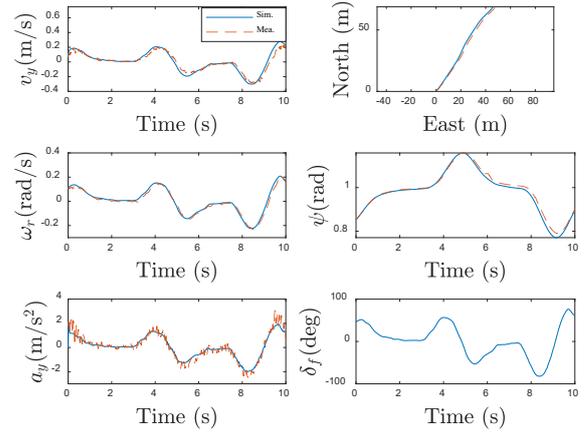

(a) longitudinal speed 30 km/h

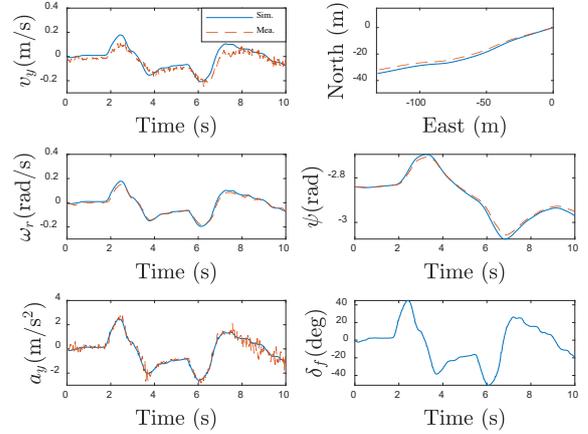

(b) longitudinal speed 50km/h

Figure 4 Comparison of model simulation results and road test measurements in doule-lane-change maneuvers at two speeds (simulation in solid blue, measurement in dashed red)

## 3. Vehicle Trajectory Prediction

### 3.1 Path Tracking Control

To follow the target trajectory, the vehicle is controlled using both the feedforward and feedback laws[24]. The feedforward part is obtained from the curvature information of the desired trajectory, and the feedback control is realized by linear quadratic regulator (LQR).

The tracking error dynamics model can be derived from the single-track vehicle model, i.e.

$$\dot{x}_{error} = Ax_{error} + B_1\delta_f + B_2\dot{\psi}_{des} \quad (5)$$

where $x_{error} = [e_1\ \dot{e}_1\ e_2\ \dot{e}_2]^T$, $e_1$ and $e_2$ denote the lateral error and heading error between the vehicle and the target path, respectively; $\dot{e}_1$ and $\dot{e}_2$ are the derivatives of $e_1$ and $e_2$ with respect to time, respectively; $\dot{\psi}_{des}$ denotes the expected yaw rate according to the path curvature $\kappa$ and the longitudinal speed $v_x$, i.e. $\dot{\psi}_{des} = v_x\kappa$. The matrices of $A$, $B_1$, and $B_2$ are as follow.

$$A = \begin{bmatrix} 0 & 1 & 0 & 0 \\ 0 & -\frac{C_f+C_r}{mv_x} & \frac{C_f+C_r}{m} & \frac{-aC_f+bC_r}{mv_x} \\ 0 & 0 & 0 & 1 \\ 0 & -\frac{aC_f-bC_r}{I_z v_x} & \frac{aC_f-bC_r}{I_z} & -\frac{a^2C_f+b^2C_r}{I_z v_x} \end{bmatrix},$$

$$B_1 = \begin{bmatrix} 0 \\ \frac{C_f}{m} \\ 0 \\ \frac{aC_f}{I_z} \end{bmatrix}, B_2 = \begin{bmatrix} 0 \\ -\frac{aC_f - bC_r}{mv_x} - v_x \\ 0 \\ -\frac{a^2C_f + b^2C_r}{I_z v_x} \end{bmatrix}.$$

The final front wheel steering control input $\delta_f$ is composed of feedforward and feedback parts, i.e. $\delta_f = \delta_f^{ff} + \delta_f^{fb}$. By solving the equilibrium of equation (5), $\dot{x}_{error}^{eq} = 0 = Ax_{error}^{eq} + B_1\delta_f^{eq} + B_2\dot{\psi}_{des}$, and setting $e_1^{eq} = 0$, we have the feedforward part as $\delta_f^{ff} = \delta_f^{eq}$.

For LQR feedback control, the cost function is set as

$$J = \frac{1}{2}\int_{t_0}^{\infty}\left[x_{error}^T W_1 x_{error} + \delta_f^2 W_2\right]dt, \quad (6)$$

where $W_1$ denotes the positive semi-definite state weight matrix; $W_2$ denotes the positive definite control weight coefficient. By solving the minimization problem of cost $J$, we have the feedback part of steering control $\delta_f^{fb} = -W_2^{-1}B_1^T P^* x_{error}$, where $P^*$ satisfies algebraic Riccati equation

$$-P^*A - A^T P^* + P^* B_1 W_2^{-1} B_1^T P^* - W_1 = 0 \quad (7)$$

Then the total control law of the front wheel steering can be finally obtained as

$$\delta_f = \delta_f^{eq} - K_{fb} x_{error} \quad (8)$$

where the feedback gain $K_{fb} = W_2^{-1}B_1^T P^*$. Note that for different longitudinal speeds $v_x$, the tracking error model varies according to vehicle dynamics equation (4). So, for different $v_x$, the corresponding gains $K_{fb}$ are obtained and stored in a look-up table for calculating the feedback steering angle.

### 3.2 EKF for Current Vehicle States

As shown in Figure 2, the on-board sensor data are firstly filtered with extended Kalman filter (EKF) to get the optimal estimation of the vehicle motion and pose information. And then a Kalman predictor is designed to make multiple predictions, so as to predict the future vehicle trajectory.

The classical Kalman filter is a recursive method, which adopts the idea of prediction and correction. The main process is as follow.
a) Prediction. The estimate of the current state and error variance are calculated to provide a prior estimate for the state at the next moment;
b) Correction. A posterior estimate is obtained by combining the prior estimate with the new measurement.

Since the model equation (4) is nonlinear, EKF is used to estimate the vehicle states. The main process is divided into the following three steps.

**Step 1**: To organize the state and measurement equations of the nonlinear system to a standard form,

$$\begin{cases} \dot{x}(t) = f(x(t), u(t)) + w(t) \\ y(t) = h(x(t)) + v(t) \end{cases} \quad (9)$$

where $x(t)$ denotes the state, $x(t) = [v_y\ \omega_r\ X_c\ Y_c\ \varphi]^T$; $u(t)$ denotes the control input, $u(t) = \delta_f$; $y(t)$ denotes the measured data from sensors; $w(t)$ and $v(t)$ denote the process noise and measurement noise, respectively.

**Step 2**: To linearize the nonlinear system model (9).

Equation (9) is expanded by Taylor expansion, and the higher-order components are abandoned. Take the partial derivatives of $f$ and $h$ with respect to the states to get the Jacobian matrices $F$ and $H$, respectively, i.e. $F = \partial f/\partial x$, and $H = \partial h/\partial x$. The continuous differential equations are then discretized and the state transition matrix is obtained as $\Phi \approx I + F \cdot t_s$, where $t_s$ is the sampling time of the algorithm.

**Step 3**: To initialize the state and error covariance matrix and then to start the filtering iterations.

The initial state and covariance matrix are

$$\hat{x}^{0|0} = \mu_0, P^{0|0} = P_0 \quad (10)$$

Using $k$ to denote the discrete time step of the system, the following iterative calculation is carried out when $k \geq 1$.

$$\begin{cases} \hat{x}^{k|k-1} = \hat{x}^{k-1|k-1} + t_s \cdot f(\hat{x}^{k-1|k-1}, u^{k-1}) \\ P^{k|k-1} = \Phi P^{k-1|k-1}\Phi^T + Q \\ K^k = P^{k|k-1}H^T\left[HP^{k|k-1}H^T + R\right]^{-1} \\ \hat{x}^{k|k} = \hat{x}^{k|k-1} + K^k\left[y^k - h(\hat{x}^{k|k-1})\right] \\ P^{k|k} = [I - K^k H]P^{k|k-1} \end{cases}$$

(11)

where $Q$ and $R$ denotes the covariance matrices of process noise $w$ and observation noise $v$, respectively. $w$ and $v$ are assumed Gaussian white noises, and satisfy the distributions of $N(0, Q)$ and $N(0, R)$, respectively.

### 3.3 Kalman Predictor with Control (KPC) for Future States

We first introduce a basic form of Kalman predictor using unchanging control inputs[25]. In order to obtain the possible vehicle trajectory in the future, the real-time estimation results of the state $\hat{x}^{k|k}$ and the error covariance matrix $P^{k|k}$ obtained in equation (11) are substituted into the multi-step Kalman predictor as follow.

$$\hat{x}^{k+i|k} = \hat{x}^{k+i-1|k} + t_s \cdot f(\hat{x}^{k+i-1|k}, u^k) \quad (12)$$

$$P^{k+i|k} = \Phi P^{k+i-1|k}\Phi^T + Q \quad (13)$$

where $\hat{x}^{k+i|k}$ denotes the predicted states at the $i^{th}$ step to predict, and $P^{k+i|k}$ is the prediction error covariance matrix at the $i^{th}$ step.

In each step of the above predictor, the control input in the state transition equation (12) uses $u^k$ at the beginning of the prediction moment $k$. It may not be a problem for a system with little change in control input by using this method. For the vehicle motion system in the lane keeping situation, the front wheel steering control $\delta_f$ may vary greatly, so using an unchanging $u^k$ in the prediction process may result in a large error of prediction result.

Fortunately, in automated driving mode, the tracking algorithm to control the front wheel steering angle can be migrated or learnt. Based on the Kalman predictor in (12) and (13), here we further add a "control prediction" procedure to predict the control input in future steps, which finally constructs Kalman Predictor with Control, as shown in Figure 5.

Using the predicted states $\hat{x}^{k+i-1|k}$, the predicted control input at the $i^{th}$ step, i.e. $\hat{u}^{k+i-1}$, can be obtained in Control Predictor procedure via known LQR control law. Then in State Prediction procedure, the vehicle states of the $i^{th}$ step, i.e. $\hat{x}^{k+i|k}$, can be calculated.

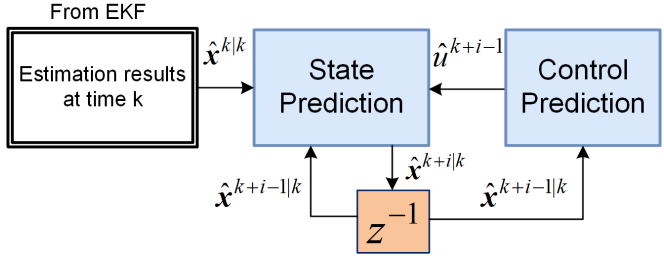

Figure 5 State prediction and control prediction in KPC

For KPC, equation (12) is changed to
$$\hat{x}^{k+i|k} = \hat{x}^{k+i-1|k} + t_s \cdot f(\hat{x}^{k+i-1|k}, \hat{u}^{k+i-1}) \quad (14)$$

With the target trajectory information, the tracking error $x_{error}$ can be expressed as $x_{error} = E(x)$, i.e.
$$\begin{cases} e_1 = E_1(X_c, Y_c) \\ e_2 = E_2(X_c, Y_c, \psi) \\ \dot{e}_1 = v_y + v_x \cdot e_2 \\ \dot{e}_2 = \omega_r - v_x \cdot \kappa \end{cases} \quad (15)$$

Then the predicted input $\hat{u}^{k+i}$ can be obtained using the control law in equation (8).
$$\hat{u}^{k+i} = \delta_f^{eq} - K_{fb} \cdot E(\hat{x}^{k+i|k}) \triangleq G(\hat{x}^{k+i|k}) \quad (16)$$

Therefore, equation (14) is revised as
$$\hat{x}^{k+i|k} = \hat{x}^{k+i-1|k} + t_s \cdot f_{cl}(\hat{x}^{k+i-1|k}) \quad (17)$$

where
$$f_{cl}(\hat{x}^{k+i-1|k}) \triangleq f\left(\hat{x}^{k+i-1|k}, G(\hat{x}^{k+i-1|k})\right)$$

Corresponding to Section 3.2, the state transition matrix $\Phi$ is changed to its closed-loop form $\Phi_{cl} = I + t_s \cdot \partial f_{cl}/\partial x$. Then equation (13) of error covariance matrix in the prediction process should also be changed to
$$P^{k+i|k} = \Phi_{cl} P^{k+i-1|k} \Phi_{cl}^T + Q \quad (18)$$

Before the Control Predictor procedure, as shown in Figure 5, we have not considered the state estimation error, which usually exists in actual process. To simulate such characteristics, the predicted tracking error $x_{error}$ is also added with a small amount of noise. A simulated noise part $v_{sim}$ is added to the predicted vehicle state $\hat{x}^{k+i|k}$, which is sampled from a stochastic process with the same statistical characteristics as the measurement noise $v(t)$ in equation (9). Then the EKF algorithm is used to filter the added noise $v_{sim}$, and we obtain the simulated estimation of $\hat{x}^{k+i|k}$, denoted as $\hat{x}_{sim}^{k+i|k}$. During prediction, we assume this estimation process reaches steady state, the Kalman filter gain $K^k$ at time $k$ is used. Finally, $\hat{x}^{k+i|k}$ in equation (16) is substituted by its simulated estimation $\hat{x}_{sim}^{k+i|k}$, and we have the final predicted control as
$$\hat{u}^{k+i} = \delta_f^{eq} - K_{fb} \cdot E(\hat{x}_{sim}^{k+i|k}) \quad (19)$$

When considering both the closed-loop system dynamics and the simulated estimation process, the error covariance matrix in equation (18) should be further improved. With the same assumption that the estimation process being in steady state, $P^{k|k}$ approximates the covariance matrix of the error between $\hat{x}_{sim}^{k+i|k}$ and $\hat{x}^{k+i|k}$, so the final form of $P^{k+i|k}$ is updated as follows.
$$P^{k+i|k} = \Phi_{cl} P^{k+i-1|k} \Phi_{cl}^T + Q$$
$$+ t_s^2 \cdot \frac{\partial f}{\partial u} K_{fb} \frac{\partial E}{\partial x} P^{k|k} \left(\frac{\partial E}{\partial x}\right)^T K_{fb}^T \left(\frac{\partial f}{\partial u}\right)^T \quad (20)$$

With above KPC procedures, we can obtain the predicted vehicle states $\hat{x}^{k+i|k}$ and the covariance matrix of prediction error $P^{k+i|k}$, which include the predicted results of vehicle CG position and body heading angle.

## 4. Lane Departure Assessment (LDA)

According to the predicted vehicle states from time $k$ and the lane boundary information, we can evaluate the risk of lane departure in the future. The evaluation of risk is mainly based on the marginal distances from the corners of the vehicle planar contour, i.e. all four rectangle corners, to the adjacent lane lines. The geometry relationship shown in Figure 1 is used to obtain the distributions of corner positions, and more details of the derivation process can be found in Appendix 2.

Then we have the Normal distribution of vehicle corner marginal distance before lane departure, as shown in Figure 6. We can design the following two rules to quantify the risk of lane departure.

Figure 6, is set to indicate when the marginal distance from corner point $j$ to its adjacent lane line, $d_j^{k+i}$, is not less than this threshold, i.e. $d_j^{k+i} \geq \Delta$, then the risk of lane departure is tolerable. Here we choose $\Delta = 0$.

(2) A probability threshold $\Pi$ is set to make sure that the predicted marginal distance $\hat{d}_j^{k+i}$ falls in the range of $[d_j^{k+i|k} - \Sigma, d_j^{k+i|k} + \Sigma]$ with a probability of $\Pi$. Here we choose $\Pi = 99.73\%$. Then the actual marginal distance satisfies
$$\Pr(\hat{d}_j^{k+i|k} - \Sigma \leq d_j^{k+i} \leq \hat{d}_j^{k+i|k} + \Sigma) = \Pi \quad (21)$$

With these two rules, if $\hat{d}_j^{k+i|k} - \Sigma \geq \Delta$, the vehicle is considered still within the allowable lateral deviation from the target trajectory at the $i^{th}$ future prediction step from time $k$, i.e. the lateral departure flag is 0. Otherwise, if $\hat{d}_j^{k+i|k} - \Sigma < \Delta$, the lateral deviation is too large, the driver should be requested to take over the driving tasks, and the lateral departure flag is 1.

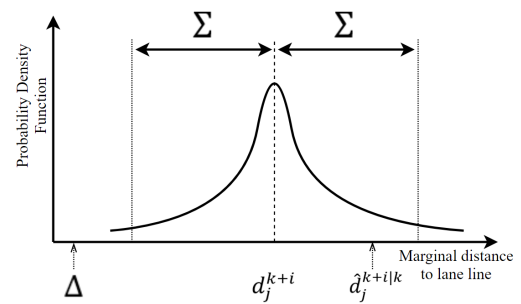

Figure 6 The distribution of the marginal distance

## 5. Validation via Simulation

### 5.1 Simulation Scenario

The joint simulation platform of Carsim and Matlab/Simulink is adopted for preliminary validation of the proposed algorithm. As shown in Figure 7, the driving scenario is designed as follow.

(1) Road: The equation of the target trajectory in the inertia coordinate system is $Y = 2$, and the lane width is 4m.

(2) Vehicle speed: The vehicle longitudinal speed is maintained near the target speed, i.e. $v_x^{des} = 30$km/h.

(3) Sensor noises: To simulate the real sensor characteristics, the zero-mean random noises with Normal distribution are injected into the vehicle state information, with the noise variances given in Table 1.

(4) Lane keeping incident: Initially, the vehicle drives at the centre of the lane and its heading is parallel to the lane line. To simulate the unknown disturbances acting on the vehicle, it deliberately deviates towards to the left in stage 1, and then in stage 2 the lane keeping controller is activated to prevent the vehicle from leaving the lane.

In the considered scenario, the vehicle is likely to deviate to the left, so the marginal distance *d* of the vehicle front left corner, as shown in Figure 1, is used to evaluate the lane departure risk.

For the benchmark algorithm, data-driven models are not considered due to their difficulties of safety validation and also fair comparison. To check the potential advantage of using a closed-loop model in prediction over open-loop models, the CTRV (Constant Turning Rate and Velocity) algorithm is selected as the benchmark algorithm. In the simulated scenarios, the longitudinal speed is assumed constant, the CTRV algorithm has satisfactory tracking and prediction performances among open-loop physics-based models[4].

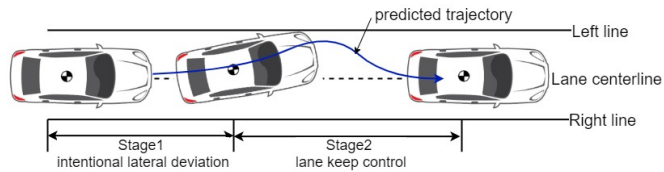

Figure 7 A simulated lane keeping incident

Table 1 Sensor noise variances in simulation

| Signal | Units | Variance |
|---|---|---|
| $v_y$ | m/s | $1 \times 10^{-6}$ |
| $\omega_r$ | rad/s | $1 \times 10^{-6}$ |
| $X_c$ | m | 1 |
| $Y_c$ | m | 1 |
| $\psi$ | rad | $1 \times 10^{-2}$ |

### 5.2 Simulation Results

The simulation process is repeated 500 times of successful lane keeping activations, with the sensor noises sampled according to Table 1. Then for both KPC and CTRV algorithms, we have trajectory prediction results and lane departure risk assessments for statistical comparisons.

Figure 8 presents an example simulation case, where the CTRV vehicle (in red) is wrongly predicted to leave the lane after 1 second, but the KPC algorithm can precisely predict the vehicle trajectory.

The overall simulation results are shown in Figure 9, where the blue circle represents the predicted position of the four vehicle corners by the KPC algorithm, the red star represents that by the CTRV algorithm, and the black solid line represents the actual trajectory of CG in the next 2 seconds.

Figure 10 shows the position prediction results of the front left corner starting from the beginning of stage 2 for every 0.1s. For each prediction step on KPC or CTRV, the points represent the results of 500 simulation cases. The orange line represents the left line of the target lane. It can be seen that the prediction results of KPC is much closer to the actual position of the front left corner than that of CTRV. With the growth of the prediction time, the prediction error of benchmark algorithm become larger, with a maximum lateral position error about 1.5m. For KPC algorithm, it can also be seen that the distribution of the predicted positions is getting more gathered with the growth of the prediction time, which means the variance of the predicted ordinates of the front left corner $\sigma_{Y_{fl}}^2$ decrease. This may be due to the existence of lane keeping algorithm, which makes the whole closed-loop vehicle system become convergent.

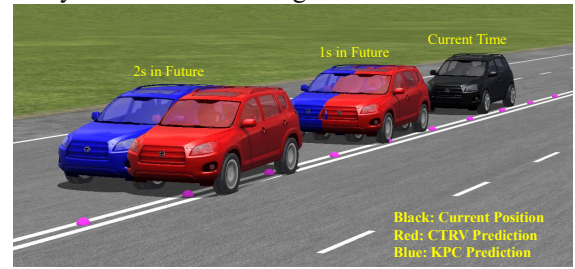

Figure 8 Prediction results of one simulation

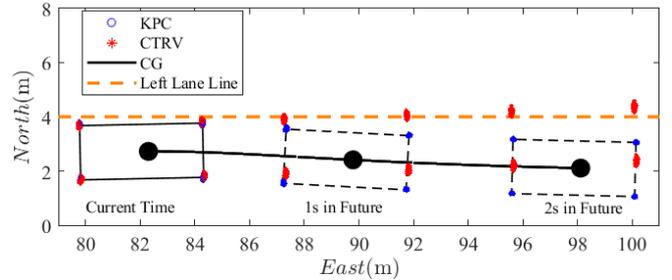

Figure 9 Overall results of all 500 simulations

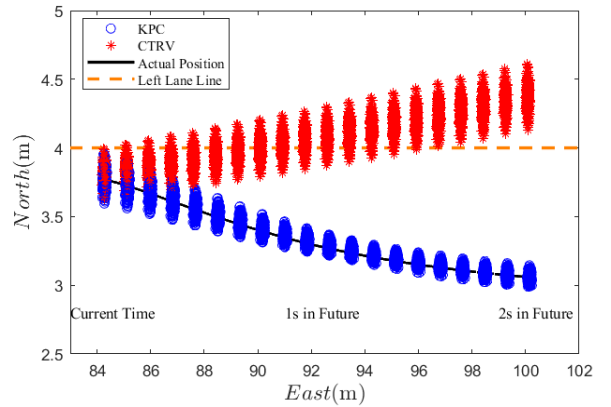

Figure 10 Front left corner prediction

Figure 10 also shows that, when calculating the covariance matrix $P^{k+i|k}$ in the process of prediction using equation (20), the variance of front left corner position $\sigma_{Y_{fl}}^2$ becomes smaller until it converges to a certain value due to the reason $\phi_1 < I$, where $I$ is the unit matrix. Figure 11 details the calculated variance $\sigma_{Y_{fl}}^2$ by KPC, the sample variance of all the prediction results $\hat{\sigma}_{Y_l}^2$ and the mean square error (MSE) of the prediction results. They are close to each other during the whole prediction process, which indicates that the calculated $\sigma_{Y_l}^2$ of KPC algorithm can describe the distribution of the errors between the predicted and the real lateral positions in a short period of time in the future.

The histograms of the predicted position distribution of the front left corner at 1s and 2s in the future are shown in Figure 12 and Figure 13, respectively. It can be seen that the predicted position at 1s and 2s follows the Normal distribution as expected.

The LDA results of both algorithms in all simulation cases are shown in Figure 14. It can be seen that in the considered cases KPC mainly believes that the vehicle is still in the lane, while CTRV is more conservative and predicts that the vehicle will drive out of the lane. The departure assessment accuracy is compared in Figure 15, which shows that with the knowledge of closed-loop vehicle behaviour, KPC works more accurately in the whole prediction process. It seems that there are some failed cases of KPC in early period of prediction, however, in these cases the vehicle is just too close to the left lane line and LDA gives wrong judgements according to a conservative threshold in equation (21).

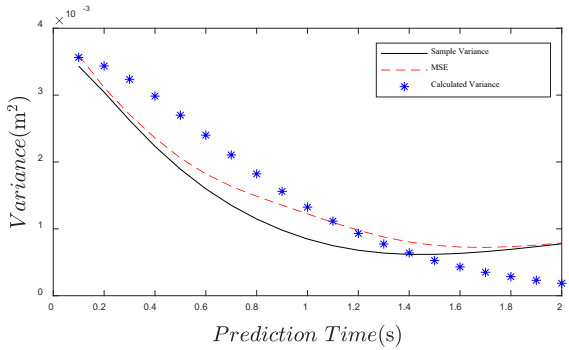

Figure 11 Lateral position variance of front left corner

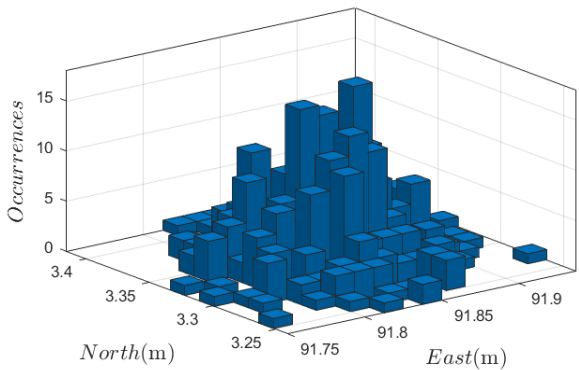

Figure 12 Histogram of predicted position at 1s

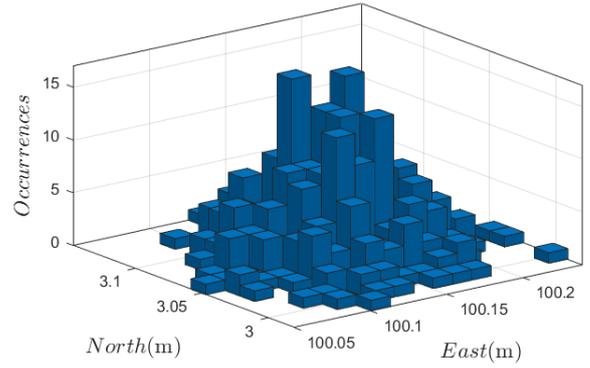

Figure 13 Histogram of predicted position at 2s

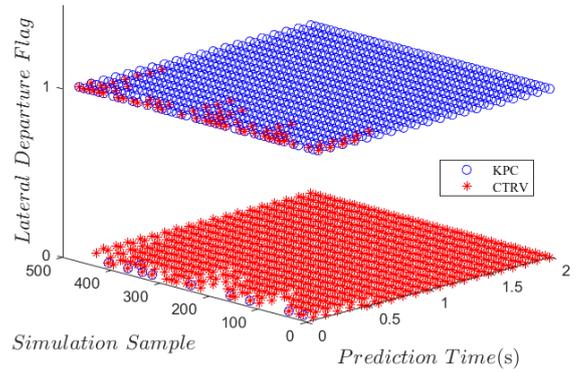

Figure 14 Lane departure assessment of both algorithms

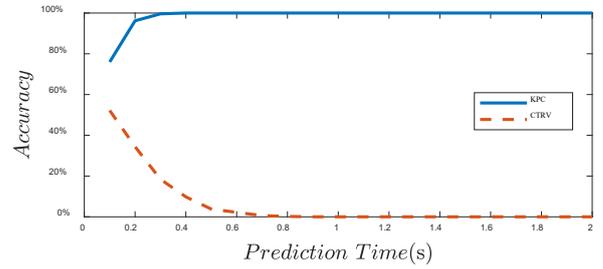

Figure 15 Departure assessment accuracy of both algorithms

## 6. Validation via On-road Test

### 6.1 Experiment setups

The test vehicle, shown in Figure 16, supports fully automated control of longitudinal and lateral motion. The on-board sensors provide main information of chassis and powertrain systems, e.g. steering wheel angle and the wheel speeds. A high precision integrated navigation system (INS), i.e. CHCNAV CGI610, is used to measure the vehicle's motion and position states. The measured signals used for algorithms are listed in Table 2.

The proposed KPC algorithm and the benchmark CTRV algorithm run in the high-performance computer at 100 Hz frequency. Based on tracking errors from sensors, the LQR-based tracking controller runs in NI CompactRIO and provides a front wheel steering angle, which is further realized in the lower-level steering-by-wire controller.

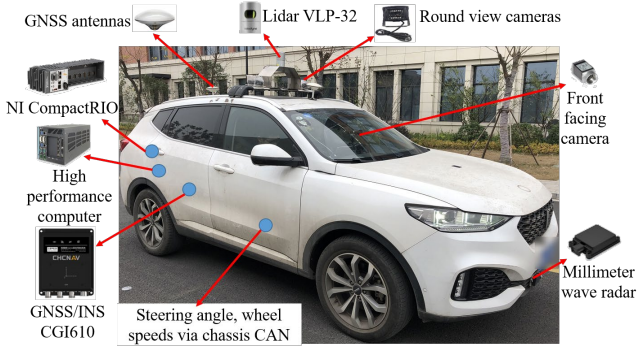

Figure 16 The test vehicle

Table 2 Measured signals in vehicle

| Signal Name | Description | Units |
|---|---|---|
| Velocity North | velocity of the measured point in the north direction | m/s |
| Velocity East | velocity of the measured point in the east direction | m/s |
| Yaw rate | vehicle yaw rate | rad/s |
| Longitude | longitude of the measured point on vehicle | deg |
| Latitude | latitude of the measured point on vehicle | deg |
| Heading angle | vehicle body heading | rad |
| Steering wheel angle | steering wheel angle | rad |
| Wheel speeds | wheel speeds in equivalent form of forward speed | m/s |

The experiment scenario and procedure are similar to that of simulation, as shown in Figure 7. During the whole process, the vehicle's longitudinal speed keeps around target speed using a PID controller. To induce lane departure incidents, in stage 1, the vehicle is deliberately steered by the test driver away from the centre of the target lane.

The experiments are divided into a low speed and a high speed groups, and are carried out in sites A and B, as shown in Figure 17. In the low speed group, the vehicle speed is 15 km/h, 20 km/h and 25 km/h, and the centre line of the target lane is a straight line of 1 deg North by East. The high-speed group includes vehicle speeds of 40 km/h and 50 km/h, and the centre line of the target lane is a straight line of 3 degrees North by West. The experiment details can be found in the video abstract attached as supplemental material.

### 6.2 Results and Discussion

The experimental results of 2 and 4 seconds prediction are shown in Figure 18 to Figure 22. In the left sub-figures, the black dot indicates the vehicle CG position when LQR controller is switched on, the solid black lines denote the vehicle trajectory after the controller is activated; the solid blue lines denote the vehicle trajectory predicted by KPC algorithm; the solid red lines denote the vehicle trajectory predicted by CTRV algorithm; the solid black rectangles denote the actual vehicle poses in the prediction period; the dashed blue rectangles denote the vehicle poses predicted by KPC algorithm; the dashed red rectangles denote the vehicle poses predicted by CTRV algorithm. The right sub-figures present the lane departure flags, in which the value 0 indicates that the vehicle is still in the lane; the value 1 indicates that the vehicle is out of the lane.

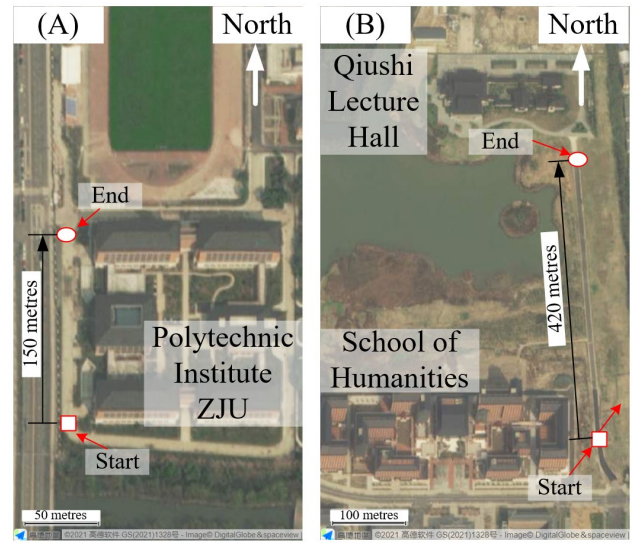

Figure 17 Test sites for low (A) and high (B) speed groups

For all cases with different speeds, KPC outperforms CTRV in both position and heading predictions. For the departure flag in all cases, CTRV predicts that the vehicle will deviate from the lane border within 1.0-2.5s, while KPC concludes correctly that the vehicle will be kept in lane. By using the additional knowledge of LQR-based closed-loop controller, with KPC the human takeover requests are not necessary to initiate too frequently, and user experience of such automated driving function can be improved.

Comparing these figures, we find that as the vehicle longitudinal speed grows from 15 to 50 km/h, it is obvious that KPC performs worse. It is because the lateral acceleration grows accordingly during lane keeping, and the linear 2DOFs model deviates from the actual vehicle dynamics. The heading angle prediction results are shown in Figure 23. The largest prediction error of heading at 15km/h is about 0.03rad (absolute value), while that at 50km/h is about 0.07rad (absolute value).

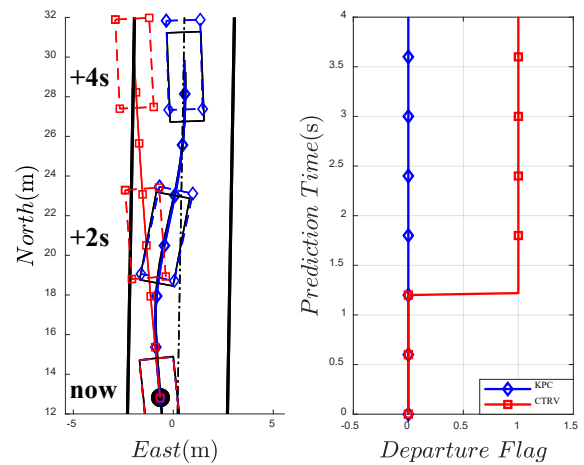

Figure 18 Trajectory and departure prediction at 15km/h

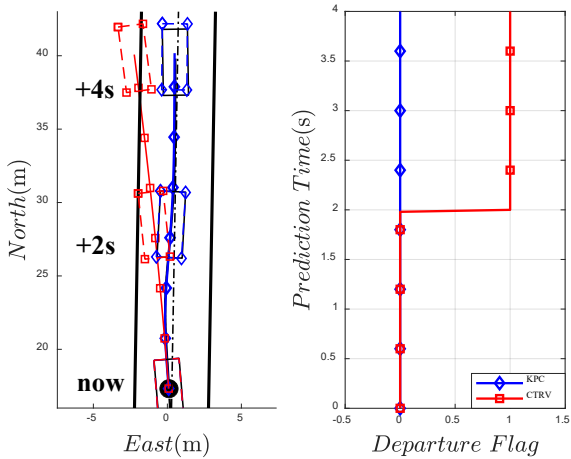

Figure 19 Trajectory and departure prediction at 20km/h

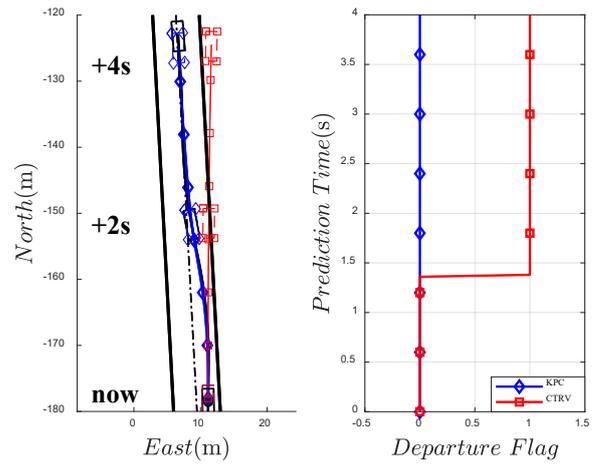

Figure 22 Trajectory and departure prediction at 50km/h

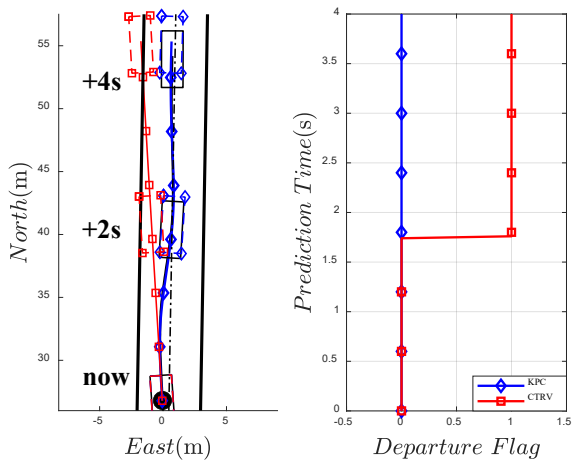

Figure 20 Trajectory and departure prediction at 25km/h

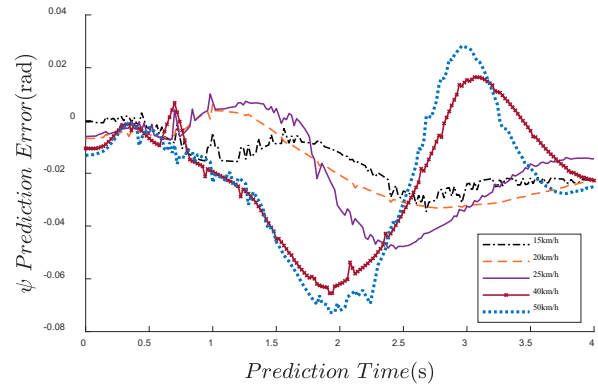

Figure 23 Heading prediction error of KPC

It can also be found that the vehicle position predicted by the two algorithms has a large longitudinal deviation from the actual vehicle position. For example, as shown in Figure 22, the longitudinal position errors in both algorithms are about 1.8m. This may be due to the fact that the actual vehicle longitudinal speed varies in the process of steering, while the prediction algorithms assume a constant longitudinal speed. However, for vehicle lane departure prediction the current assumption of constant speed still works. Considering that the prediction performance may get deteriorated when the vehicle longitudinal speed varies greatly, a separate speed prediction can compensate for this disadvantage. For better prediction, a similar KPC methodology can also be adopted to include the longitudinal motion prediction.

## 7. Conclusion

A lane departure prediction method based on closed-loop vehicle dynamics is proposed, and it's applied in human takeover decision for automated lane keeping scenario. It is an improved model of physics-based prediction, and can provide both high computing efficiency and long-term prediction accuracy. Simulation results show the algorithm is effective in constant speed lane keeping scenarios, and it can describe the probability distribution characteristics of vehicle pose prediction. Experiment results show that the vehicle trajectory prediction algorithm is effective at speeds of 15 to 50 km/h. It may work as a promising solution to lane

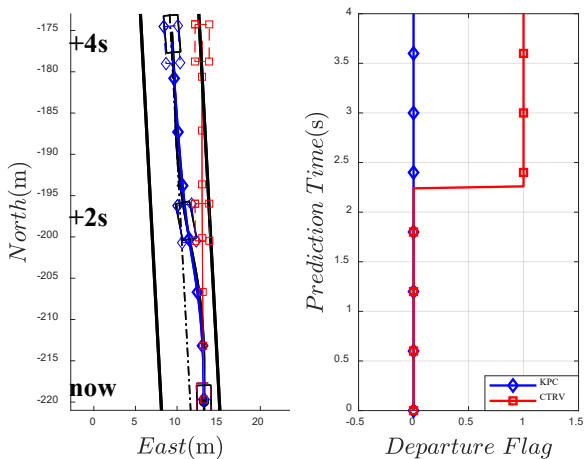

Figure 21 Trajectory and departure prediction at 40km/h

departure risk assessment for automated lane keeping functions.

However, there are also some limitations of this article, which can be further extended in future work.

(1) The effects of unknown disturbances and vehicle control actuator errors are not considered in the prediction algorithm, which are common in practical vehicle control. For improvements, a specific observer helps to predict sidewinds, and more accurate actuator modelling can improve the prediction, too. Online estimation of system parameters can also contribute to better model accuracy.

(2) For demonstration, here the control law in the predictive algorithm is directly migrated from the actual controller, assuming the developer can access the source codes of lane keeping controller. In some cases, the detailed law of the tracking controller may be unknown to the developer, so it is necessary to design an algorithm to learn the law from controller inputs-outputs mapping. In view of this, some data-driven algorithms can be used to learn the actual control law in the future work.

(3) This study focuses on the lateral dynamics of the vehicle, while the longitudinal vehicle dynamics is not considered. In future work, more complex scenarios can be considered, where the longitudinal and lateral models of vehicles should be integrated. Additionally, the algorithm reliability for human takeover in dangerous scenarios can be further validated using simulated driving or more comprehensive real road tests.

**Supplementary materials**

A supplementary video abstract is provided, which includes a brief introduction of the proposed approach and an example of validation experiments. https://youtu.be/rRfUUpBDxV8

**Acknowledgements**

The authors appreciated Anfei Zha for his help in carrying out the real vehicle experiments, and Linhui Chen for his help in preparing the video abstract.

**Declaration of conflicting interests**

The author(s) declared no potential conflicts of interest with respect to the research, authorship, and/or publication of this article.

**Funding**

The author(s) disclosed receipt of the following financial support for the research, authorship, and/or publication of this article: This work was supported by Department of Science and Technology of Zhejiang (No. 2022C01241, No. 2018C01058), and Ningbo Science & Technology Bureau (No. 2018B10063, 2018B10064).

**ORCID iDs**

Daofei Li, https://orcid.org/0000-0002-6909-0169
Siyuan Lin, https://orcid.org/0000-0001-5161-5652
Guanming Liu, https://orcid.org/0000-0002-2073-0460

**Appendix 1: Nomenclature**

| Variable | Description |
|---|---|
| $a, b$ | Distances between the centre of mass and the front/rear axles, 1.13m and 1.55m |
| $B$ | Half width of vehicle, 0.93m |
| $C_{f,r}$ | Lateral stiffness of the front and rear axle tyres, 1e5 N/rad and 2e5 N/rad |
| $d_j^{k+i}$ | Actual value of the distance between corner point $j$ and its adjacent lane line at time $k+i$ |
| $\hat{d}_j(k+i\|k)$ | Predicted value of the distance between corner point $j$ and its adjacent lane line at time $k+i$ |
| $e_1, e_2$ | Lateral error and heading error between vehicle and target trajectory |
| $F_{yf,yr}$ | Side forces of front and rear tyres |
| $F, H$ | Jacobian matrices of $f$ and $h$ |
| $i$ | Prediction step, $i^{th}$ step |
| $I_z$ | Vehicle yaw moment of inertia, $3200\text{kg}\cdot\text{m}^2$ |
| $k$ | Discrete time, $k^{th}$ step |
| $K_{fb}$ | Feedback gain |
| $l_f$ | Distance between vehicle's centre of mass and the front end of vehicle contour, 2.11m |
| $m$ | Vehicle mass, 2030kg |
| $Q, R$ | Covariance matrices of process noise and observation noise |
| $t_s$ | Sampling time of the algorithm |
| $u(t)$ | Control inputs |
| $v_x$ | Longitudinal velocity of vehicle CG |
| $v_y$ | Lateral velocity of vehicle CG |
| $v(t), w(t)$ | Measurement and process noises |
| $W_1, W_2$ | State weighting matrix and control weighting |
| $x(t)$ | System states |
| $y(t)$ | Measured outputs |
| $X_c, Y_c$ | East and North coordinates of CG in the inertia coordinate system |
| $X_c^{k+i}, Y_c^{k+i}$ | Actual East and North coordinates of vehicle's centre of mass at time $k+i$ |
| $X_{left}, Y_{left}$ | Coordinates of front left corner |
| $X_{right}, Y_{right}$ | Coordinates of front right corner |
| $\alpha_{f,r}$ | Slip angles of front and rear tyres |
| $\delta_f$ | Vehicle front wheel steering angle |
| $\kappa$ | Curvature of target path |
| $\Phi$ | State transition matrix |
| $\Phi_{cl}$ | Modified state transition matrix |
| $\psi$ | Vehicle heading angle |
| $\psi^{k+i}$ | Actual value of heading angle of vehicle at time $k+i$ |
| $\dot{\psi}_{des}$ | Desired yaw rate |
| $\omega_r$ | Vehicle yaw rate |
| Subscripts $j = fl, fr, rl, rr$ | Front left, front right, rear left, rear right |

**Appendix 2: Derivation for the distribution of marginal distance d**

Based on the probabilistic distribution of CG position and vehicle heading, the task is to determine the probabilistic distribution of the marginal distance of all four corners to their corresponding lane lines.

Assuming that the vehicle shape is a rectangle when viewed from top, as shown in Figure 1, we use $l_f$ to denote the distance from CG to the front end of the vehicle contour, and $B$ to denote the vehicle half width. Taking the front left corner as an example, its coordinates are

$$\begin{cases} X_{fl} = X_c + l_{fB} \cdot \cos(\psi + \varphi) \\ Y_{fl} = Y_c + l_{fB} \cdot \sin(\psi + \varphi) \end{cases} \quad (22)$$

where $l_{fB} = \left(l_f^2 + B^2\right)^{1/2}$, and $\varphi = \operatorname{atan}(B/l_f)$.

The predicted vehicle heading angle includes two parts, i.e. the actual value $\psi$ and the predicted error $\Delta\psi$, $\hat{\psi} = \psi + \Delta\psi$. According to Taylor expansion, the nonlinear part containing trigonometric functions can be expanded as

$$\begin{cases} \cos(\hat{\psi} \pm \varphi) \approx \cos(\psi \pm \varphi) - \sin(\psi \pm \varphi) \cdot \Delta\psi \\ \sin(\hat{\psi} \pm \varphi) \approx \sin(\psi \pm \varphi) + \cos(\psi \pm \varphi) \cdot \Delta\psi \end{cases} \quad (23)$$

According to the prediction algorithm, the prediction result of step $i$ is shown in Table 3.

Table 3 The elements of $\hat{x}^{k+i|k}$ and $P^{k+i|k}$

| Predicted vehicle states $\hat{x}^{k+i|k}$ | Prediction error covariance matrix $P^{k+i|k}$ |
|---|---|
| $\begin{bmatrix} \hat{v}_y^{k+i\|k} \\ \hat{\omega}_r^{k+i\|k} \\ \hat{X}_c^{k+i\|k} \\ \hat{Y}_c^{k+i\|k} \\ \hat{\psi}^{k+i\|k} \end{bmatrix}$ | $\begin{bmatrix} \sigma_{v_y}^2 & cov(v_y,\omega_r) & cov(v_y,X_c) & cov(v_y,Y_c) & cov(v_y,\psi) \\ cov(v_y,\omega_r) & \sigma_{\omega_r}^2 & cov(\omega_r,X_c) & cov(\omega_r,Y_c) & cov(\omega_r,\psi) \\ cov(v_y,X_c) & cov(\omega_r,X_c) & \sigma_{X_c}^2 & cov(X_c,Y_c) & cov(X_c,\psi) \\ cov(v_y,Y_c) & cov(\omega_r,Y_c) & cov(X_c,Y_c) & \sigma_{Y_c}^2 & cov(Y_c,\psi) \\ cov(v_y,\theta) & cov(\omega_r,\theta) & cov(X_c,\theta) & cov(Y_c,\theta) & \sigma_\psi^2 \end{bmatrix}$ |

For simplicity, the covariances between vehicle states are ignored, which is appropriate in this short-term prediction problem. Then the vehicle coordinates, $\hat{X}_c^{k+i|k}$, $\hat{Y}_c^{k+i|k}$ and the heading angle $\hat{\psi}^{k+i|k}$, conform to independent Normal distributions as follow,

$$\begin{cases} \hat{X}_c^{k+i|k} \sim N(X_c^{k+i}, \sigma_{X_c}^2) \\ \hat{Y}_c^{k+i|k} \sim N(Y_c^{k+i}, \sigma_{Y_c}^2) \\ \hat{\psi}^{k+i|k} \sim N(\psi^{k+i}, \sigma_\psi^2) \end{cases} \quad (24)$$

where $X_c^{k+i}$ and $Y_c^{k+i}$ denote the actual value of the abscissa and ordinate of the vehicle's centre of mass at time $k+i$, and $\psi^{k+i}$ denotes the actual value of the heading angle of vehicle at time $k+i$.

For the prediction process of the $i^{\text{th}}$ step, the coordinates of the predicted position of the front left corner, $(\hat{X}_{fl}^{k+i|k}, \hat{Y}_{fl}^{k+i|k})$, also approximately follow the Normal distribution, according to Equations (22)-(24).

$$\begin{cases} \hat{X}_{fl}^{k+i|k} \sim N(X_c^{k+i} + l_{fB}\cos(\psi_+^{k+i}), \sigma_{X_c}^2 + l_{fB}^2 \sin^2(\psi_+^{k+i}) \sigma_\psi^2) \\ \hat{Y}_{fl}^{k+i|k} \sim N(Y_c^{k+i} + l_{fB}\sin(\psi_+^{k+i}), \sigma_{Y_c}^2 + l_{fB}^2 \cos^2(\psi_+^{k+i}) \sigma_\psi^2) \end{cases} \quad (25)$$

where $\psi_+^{k+i} = \psi^{k+i} + \varphi$.

Assuming that the actual marginal distance from corner point $j$ to its adjacent lane line can be expressed as a function $d_j = D_j(X_j, Y_j)$, the predicted value of $d_j$ is $\hat{d}_j^{k+i|k} = D_j(\hat{X}_j^{k+i|k}, \hat{Y}_j^{k+i|k})$, which can be further expanded as follows using Taylor formula,

$$\hat{d}_j^{k+i|k} \approx D_j\left(X_j(k+i|k), Y_j(k+i|k)\right) + \frac{\partial D_j}{\partial X_j} \cdot \Delta X_j + \frac{\partial D_j}{\partial Y_j} \cdot \Delta Y_j \quad (26)$$

Then the marginal distance $d_j$ can be described using a Normal distribution,

$$\hat{d}_j^{k+i|k} \sim N\left(d_j^{k+i}, (\frac{\partial D_j}{\partial X_j})^2 \sigma_{X_j}^2 + (\frac{\partial D_j}{\partial Y_j})^2 \sigma_{Y_j}^2\right) \quad (27)$$